\documentclass[journal]{IEEEtran}
\usepackage[caption=false,font=normalsize,labelfont=sf,textfont=sf]{subfig}
\usepackage[pdftex]{graphicx}
\usepackage{stfloats}
\usepackage{rotating}
\usepackage{graphicx}  
\usepackage{epstopdf}
\usepackage{amsmath}
\usepackage{amssymb}
\usepackage{textcomp,booktabs}
\usepackage{multirow}
\usepackage[usenames]{color}
\usepackage{wrapfig}
\usepackage{longtable}
\usepackage{colortbl,booktabs}
\usepackage{verbatim}
\usepackage{caption}
\usepackage{cite}
\usepackage[colorlinks,linkcolor=blue]{hyperref}

\usepackage{bigstrut}

\begin{document}
%

\title{Data-Level Recombination and Lightweight Fusion Scheme for RGB-D Salient Object Detection}


\author{Xuehao Wang$^{1}$ ~~~~Shuai Li$^{1}$
 ~~~~Chenglizhao Chen$^{1,2*}$\thanks{Corresponding author: Chenglizhao Chen, cclz123@163.com.}~~~Yuming Fang$^{3}$~~~Aimin Hao$^{1}$~~~Hong Qin$^4$\\
$^1$Qingdao Research Institute \& State Key Laboratory of VRTS, Beihang University\\
$^2$Qingdao University~~~
$^3$Jiangxi University of Finance and Economics~~~$^4$Stony Brook University\\
Code\&Data, \url{https://github.com/XueHaoWang-Beijing/DRLF}}

\markboth{IEEE Transactions on Image Processing, VOL.XX, NO.XX, XXX.XXXX}%
{Shell \MakeLowercase{\textit{et al.}}: Bare Demo of IEEEtran.cls for Journals}

\maketitle

\IEEEtitleabstractindextext{
\begin{abstract}
Existing RGB-D salient object detection methods treat depth information as an independent component to complement its RGB part, and widely follow the bi-stream parallel network architecture.
To selectively fuse the CNNs features extracted from both RGB and depth as a final result, the state-of-the-art (SOTA) bi-stream networks usually consist of two independent subbranches; i.e., one subbranch is used for RGB saliency and the other aims for depth saliency.
However, its depth saliency is persistently inferior to the RGB saliency because the RGB component is intrinsically more informative than the depth component.
The bi-stream architecture easily biases its subsequent fusion procedure to the RGB subbranch, leading to a performance bottleneck.
In this paper, we propose a novel data-level recombination strategy to fuse RGB with D (depth) before deep feature extraction, where we cyclically convert the original 4-dimensional RGB-D into \textbf{D}GB, R\textbf{D}B and RG\textbf{D}.
Then, a newly lightweight designed triple-stream network is applied over these novel formulated data to achieve an optimal channel-wise complementary fusion status between the RGB and D, achieving a new SOTA performance.
\end{abstract}


}
\maketitle
\IEEEdisplaynontitleabstractindextext
\IEEEpeerreviewmaketitle

\section{Introduction and Motivation}
\IEEEPARstart{S}alient object detection (SOD) aims to distinguish the most visually distinctive objects from non-salient surroundings nearby~\cite{CVPR_H2017,fang2019visual,ISMAR2020}.
As a widely used preprocessing tool, the SOD related down-stream applications include various computer vision tasks, such as object detection~\cite{sun2019feature}, image expression~\cite{battiato2014saliency}, image retrieval~\cite{zhang2012conjunctive,liu2013model}, image compression~\cite{TIP_C2010,CME_J2017}, image retargeting~\cite{fang2012saliency,yan2017codebook}, visual tracking~\cite{CC2015PR,CC2016PR}, video saliency~\cite{chen2019improved,cong2019video,CC2019TMM2,OurTIP17} and video segmentation~\cite{fan2019shifting,wang2019semi,zeng2019joint,wang2017video}.

\begin{figure}[t]
	\begin{center}
		\includegraphics[width=1.0\linewidth]{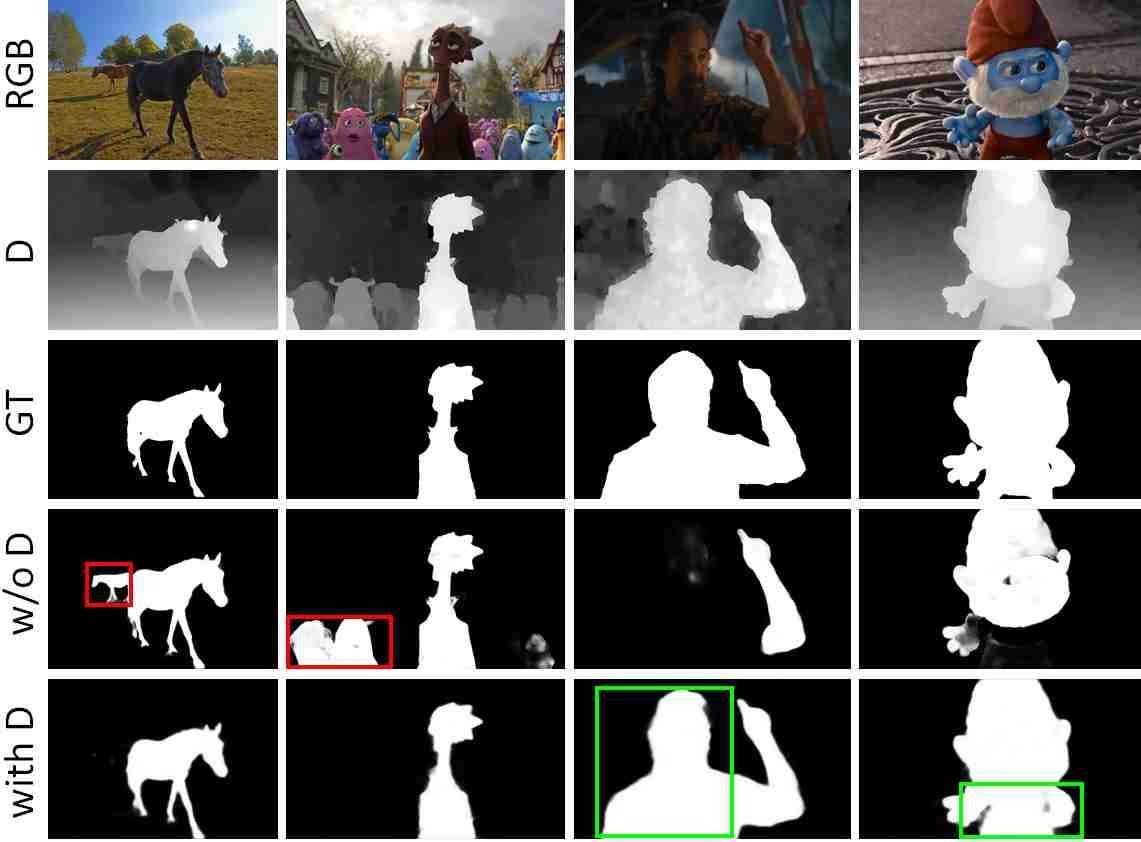}
	\end{center}
	\vspace{-0.2cm}
	\caption{Qualitative demonstrations towards the contribution of the depth (D) information.
We use red boxes to highlight those false-alarm detections of the approach without using D (i.e., ``w/o D'').
Benefited from D, those regions originally miss-detected by the ``w/o D'' approach now can be detected correctly and we use green boxes to highlight them.}
	\label{fig:Demo_RGBD}
	\vspace{-0.4cm}
\end{figure}

Different to the conventional salient object detection methods~\cite{kong2018exemplar,CC2019TMM1,zhang2020multistage,CC2015TIP} using RGB information solely, the RGB-D salient object detection methods~\cite{TIP_Q2017,piao2019saliency,chen2020improved} have achieved significant performance improvements due to the newly available depth information (see Fig.~\ref{fig:Demo_RGBD}).
In fact, the RGB component is intrinsically more informative than the depth component, which is abundant in texture and color information, and it should exhibit better performance than the depth component.
Nevertheless, the fact may occasionally be the opposite, in which we may easily obtain a high-quality saliency map over the depth channel if it satisfies the following aspects: \underline{1)} the depth information is correctly sensed by the depth-sensing equipment; \underline{2)} the salient object is located at a different depth layer to its non-salient surroundings nearby.

\begin{figure}[t]
  \centering
  \includegraphics[width=1\linewidth]{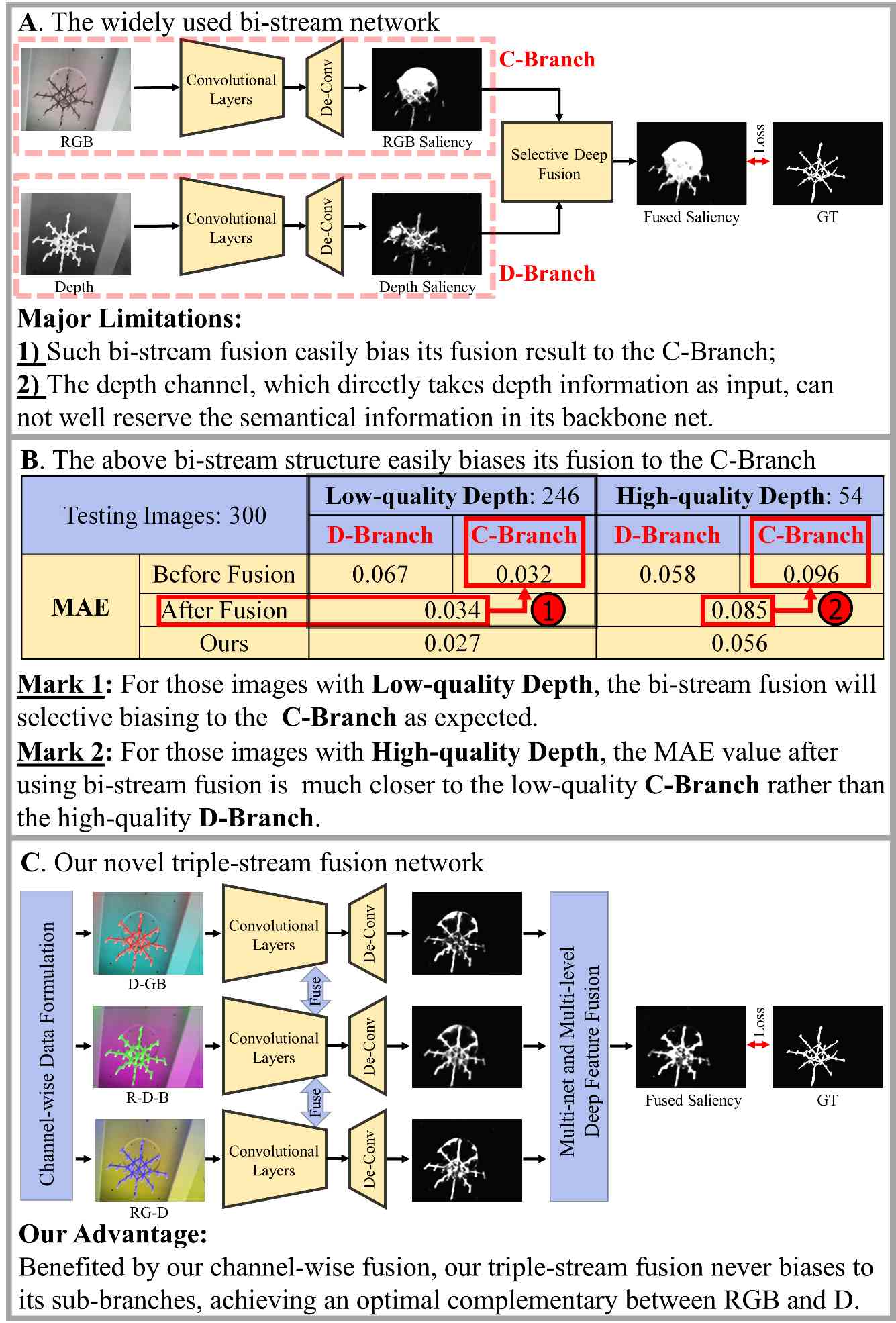}
  \caption{The main highlight of this paper. Though the conventional bi-stream fusion will achieve better performance than either subbranch, its fused result easily biases to the RGB channel, failing to make full use of the depth information, see the quantitative results in subfigure B (more quantitative results can be found in the component evaluation section).
  Meanwhile, we have listed the advantages of our newly proposed triple stream fusion network.
  All deep models mentioned are trained over an \textbf{identical} training set containing both high-quality and low-quality cases.  We divide the testing set into the “high-quality” subset and the “low-quality” subset to highlight the biased fusion status of the conventional bi-stream networks.}
  \label{fig:motivation}
  \vspace{-0.4cm}
\end{figure}

The conventional SOTA RGB-D salient object detection methods~\cite{wang2019adaptive} have adopted the bi-stream network architecture to pursue a complementary status between RGB and depth, in which one of its streams is used for RGB saliency prediction and the other aims for depth saliency calculation.
The final salient object detection results are obtained by feeding the previous RGB/depth saliency maps into a selective fusion module, and we show its pipeline in Fig.~\ref{fig:motivation}A.

However, there exists one major problem which seems to lead the classic bi-stream architecture based methods to reach a performance bottleneck; i.e., the depth quality usually differs from scene to scene, and the strong data adaptability of the current deep learning based techniques may lead the trained fusion subnet to bias towards the informative RGB component, producing a mediocre detection even in the case that the depth channel is trustworthy, and the quantitative results can be found in Fig.~\ref{fig:motivation}B.

To solve the aforementioned problem, we propose a channel-wise fusion scheme to integrate depth channel into RGB component, in which we use depth (D) channel to cyclically replace each sub-channel of RGB, obtaining 3 independent 3-dimensional data; i.e., \textbf{D}GB, R\textbf{D}B and RG\textbf{D}, which can respectively be fed into any off-the-shelf RGB salient object detection model (e.g., \cite{CVPR_H2017}) to produce a much improved saliency estimation than the solely depth channel based one. The effectiveness of the proposed channel-wise fusion scheme is proved in our ablation experiments.
Meanwhile, to take full advantage of the novel data, we design our deep network as a triple-stream architecture, in which each subbranch will receive 1 of the 3 fused data as input to avoid the bias problem.
Moreover, to ensure an optimal complementary status between RGB and D, we propose a lightweight recursive fusion strategy to interactively complement each subbranch of our network.

Also, we have conducted an extensive quantitative evaluation to validate the effectiveness of our method, in which we have compared our method with 16 SOTA methods over 5 widely adopted benchmark datasets.
Overall, the main contributions of this paper can be summarized as:
\begin{itemize}
\item We have designed a novel data-level fusion scheme to integrate the RGB component with the depth channel, ensuring the high-quality and low-level saliency estimations;
\item We have proposed a novel lightweight triple-stream fusion network to make full use of our newly formulated input data, ensuring an optimal complementary fusion status between RGB and D;
\item We have conducted extensive validations and comparisons to show the effectiveness and advantages of our method, and our code is also public available.
\end{itemize}

\begin{figure*}[t]
  \centering
  \includegraphics[width=1\linewidth]{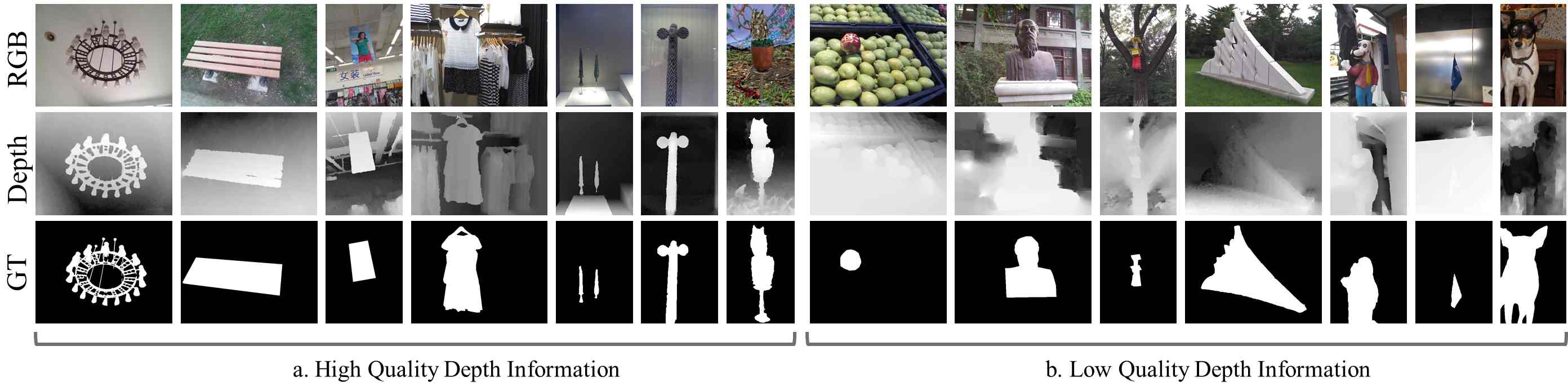}
  \caption{Demonstration of the depth quality, in which the left component shows the high-quality cases, while the right component shows the low-quality cases. GT represents the human well-annotated binary saliency ground truth.}
  \vspace{-0.2cm}
  \label{FigureDquality}
  \end{figure*}

\section{Related Works}
In general, the depth-related saliency estimation is motivated by an assumption that the salient object should be located at a different depth layer to its non-salient surroundings nearby.
Thus, similar to the conventional RGB salient object detection methods, the key rationale for saliency estimation over the depth channel is conducting multi-level/multi-scale contrast computations.

\vspace{-0.2cm}
\subsection{Hand-Crafted Methods}
As one of the most representative hand-crafted methods, Peng et al.~\cite{ECCV_P2014} adopted the multi-level (e.g., local, global and background) contrast computation to obtain multi-contextual depth saliency.
Similarly, Ren et al.~\cite{CVPRW_J2015} devised a regional contrast to further boost the depth saliency quality.
By using the off-the-shelf PageRank technic, the previously computed depth saliency features were fused with multiple RGB global priors as the final detection results.
Although the regional contrast computation can really robust detection results in RGB-D images with simple backgrounds, it may occasionally produce massive false-alarm detections with cluttered backgrounds.
To alleviate it, Feng et al.~\cite{CVPR_F2016} proposed a region-wise angular contrast computation over depth information, in which the angular contrast degree, as an alternate saliency clue, may potentially be able to separate the salient object from its non-salient surroundings nearby.
Once the hand-crafted depth saliency was computed, these conventional methods would frequently follow the linear fusion scheme to integrate RGB saliency with depth saliency for an improved salient object detection result, in which the fusion weight can either be empirically assigned or be adaptively formulated~\cite{Cong2016Saliency,cong2017iterative}.

\vspace{-0.2cm}
\subsection{Deep Learning based Methods}
After entering the deep learning era, the SOTA methods have widely adopted the deep learning based technics for a high-performance RGB-D salient object detection.
Qu et al.~\cite{TIP_Q2017} utilized the compactness prior to guide the saliency detection over depth information, which would then be fused with its RGB saliency by using the CNNs based selective fusion module.
However, the result may occasionally encounter various detection artifacts, because both its RGB saliency and depth saliency were computed in a hand-crafted manner.
To solve it, Shigematsu et al.~\cite{ICCV_S2017} proposed a bi-stream CNNs network to extract patch-wise deep features from RGB and depth respectively.
Then, both the RGB and depth-based deep features were concatenated and latterly fed into multiple fully connected layers to achieve a selective deep fusion.

Most recently, the FCNs network has been widely adopted to conduct an end-to-end RGB-D salient object detection.
Zhu et al.~\cite{Zhu2018PDNet} followed the bi-stream network architecture, in which one of its subnets aimed to conduct FCNs based RGB saliency estimation, and the other focused on depth saliency revealing.
Meanwhile, its subsequent RGB-D fusion module was also quite simple, which directly convolved the deep features generated by its depth and RGB subbranches to achieve an improved RGB salient object detection result.
Although the depth features can really benefit the RGB saliency estimation in most cases, it may occasionally encounter failure detections when the depth information was less trustworthy (e.g., low-quality depth, see the right component of Fig.~\ref{FigureDquality}), leading the fused result even worse.
Thus, Wang et al.~\cite{wang2019adaptive} weakly learned an additional fusion indicator, which was capable of adaptively ensuring its fusion procedure to bias towards its RGB component if the depth channel quality was predicted to be less trustworthy.
Further, Zhao et al.~\cite{zhao2019contrast} used an additional contrast loss to enhance the depth quality, achieving a much improved SOD result.
Although many improvements have been made, the aforementioned methods still compute their depth saliency using depth information solely, which inevitably bias the fusion procedure towards the RGB component.

\section{The Proposed Method}
The semantic information usage is a vital factor to determine the overall SOD performance, and thus the deep learning based SOTA methods have widely adopted the off-the-shelf backbones (e.g., VGG16) to compute their high-discriminative deep features.
Since most of these prevailing backbones are pre-trained using large-scale ``3-dimensional'' training set with strong semantic knowledge (e.g., RGB images with human-labeled semantic categories), the deep features from these backbones are frequently embedded with strong semantic information even if these backbones are fine-tuned over other 3-dimensional SOD training sets.

In the face of 4-dimensional RGB-D data, we need to completely retrain these backbones if we want to continue taking full advantage of the semantic information embedded in the pre-trained feature backbones.
However, to the best of our knowledge, there exists no large-scale 4-dimensional RGB-D training set with strong semantic information.
Thus, the existing RGB-D SOD approaches have considered the bi-stream network structure, including one RGB saliency branch, which takes 3-dimensional RGB data as input as usual, and one D saliency branch which duplicates depth channel 2 times as ``DDD''.
And these individual saliency subbranches will latterly be fused selectively to output the final RGB-D saliency maps.

Though this methodology is reasonable and effective in most cases, it has reached a performance bottleneck, because this methodology has overlooked one critical fact, i.e., the strong data adaptability of current deep learning based techniques may lead the selective fusion process to bias towards one of its preceding subbranches ``extremely'' if there exists a significant performance gap between these subbranches.
To be more specific, as we have mentioned before, the RGB saliency subbranch usually outperforms the D saliency subbranch significantly, and, as a result, the valuable depth information, which is supposed to be capable of benefiting the SOD task, may be overwhelmed during the bi-stream selective deep fusion---a biased fusion process.

To solve it, here we propose a simple yet effective way to channel-wisely fuse depth information with RGB information, which is capable of making full use of both RGB and depth information, and avoiding the abovementioned unbalanced fusion result.

\begin{figure*}[t]
  \centering
  \includegraphics[width=0.9\linewidth]{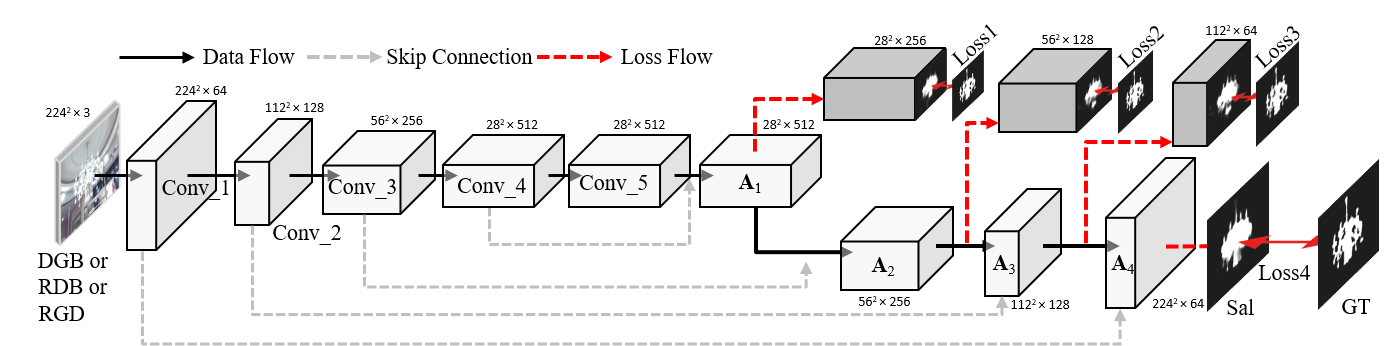}
  \centering
  \caption{Architecture of our backbone subnet.}
  \vspace{-0.4cm}
  \centering
  \label{fig:backbone}
\end{figure*}

\subsection{Channel-Wise Data Fusion}
The conventional methods compute their depth saliency using depth information solely, which easily lead to worse depth saliency because the 3-dimensional ``DDD'' is still quite different to the ``RGB'' data---a large data gap in fact, degenerating their feature backbones in providing semantic deep features.

As an independent channel to separate salient objects from their non-salient surroundings nearby, the nature of D channel is identical to RGB channels, and thus it is intuitive and reasonable to recombine these different channels into new data formulations to diminish such ``data gap''.

Our novel idea is to insert D channel into RGB channels cyclically.
That is, we re-formulate the original 4-dimensional RGB-D data into 3 independent 3-dimensional data, i.e., \textbf{D}GB, R\textbf{D}B and RG\textbf{D}, in which we use the depth information to replace one channel of RGB data each time.
Consequently, such new data are exactly within a 3-dimensional formulation, and, compared with the conventional ``RGB+DDD'', the feature backbones can output deep features with more semantic information after being fed by these new data.

\subsection{Novel Backbone Network}
So far, we can directly use the off-the-shelf pre-trained network (i.e., we simply choose the vanilla VGG16) to extract high-dimensional deep features respectively for each D-GB, R-D-B and RG-D, and we show the detailed network architecture in Fig.~\ref{fig:backbone}.

In our implementation, we have dropped the VGG16's last pool layer and the fully connected layers.
The outputs of five convolutional blocks (separated by pool layers) are denoted as $\mathrm{Conv\_i, i\in\{1,2,3,4,5\}}$.
Meanwhile, we have modified the hyperparameters of $\mathrm{Pool\_4}$ to ensure abundant tiny details in $\mathrm{Conv\_5}$; i.e., kernel size, padding number and stride in $\mathrm{Pool\_4}$ layer are assigned to \{3, 1, 1\} respectively.

To make full use of the multi-level deep features, we also recursively combine deep features from consecutive levels by feeding them into 2 convolutional layers with 1 up-sampling layer for skip connections.
We denote the deep feature of each skip connection as $\textbf{A}_i$($i \in [1,4]$), which is supervised by ground truth separately to ensure the robustness of our network.
Moreover, we normalize the $\textbf{A}_i$ before each skip connection.
Thus, the outputs of multi-level skip connections can be formulated by Eq.~\ref{eq:feature_A}.
\begin{equation}
\label{eq:feature_A}
\mathrm{\textbf{A}_i}=\left\{\begin{array}{ll}{f\big(bn(\mathrm{Conv\_j}) \circ bn(\textbf{A}_{i-1})\big)} & {i \in [2,4]} \\ \\ {f\big(bn(\mathrm{Conv\_4}) \circ bn(\mathrm{Conv\_5})\big)} & {i = 1}\end{array}\right.,
\end{equation}
where $\circ$ denotes the feature concatenation operation; $\textbf{A}_i$ denotes the feature computed by the $i$-th skip connection; we set $j=5-i$, $bn$ denotes the batch normalization function; $f$ denotes a $3\times3$ convolutional operation, which transform its input data into specific channel numbers as described in Fig.~\ref{fig:backbone}.
Here we use $\rm Sal$ to represent the deepest data flow in our backbone subnet (see Eq.~\ref{eq:saliency-branch}).
\begin{equation}
\label{eq:saliency-branch}
\begin{aligned}
&\mathrm{Sal} =\\
&f\Bigg(bn(\textbf{A}_4) \circ bn\Big(f\Big(bn(\textbf{A}_3) \circ bn\big(f\big(bn(\textbf{A}_2) \circ bn(\textbf{A}_1)\big)\big)\Big)\Big)\Bigg).
\end{aligned}
\end{equation}

\subsection{Multi-Level and Multi-Net Deep Feature Fusion}
\label{Sec:fusion}
\begin{figure}[h]
  \centering
  \includegraphics[width=0.8\linewidth]{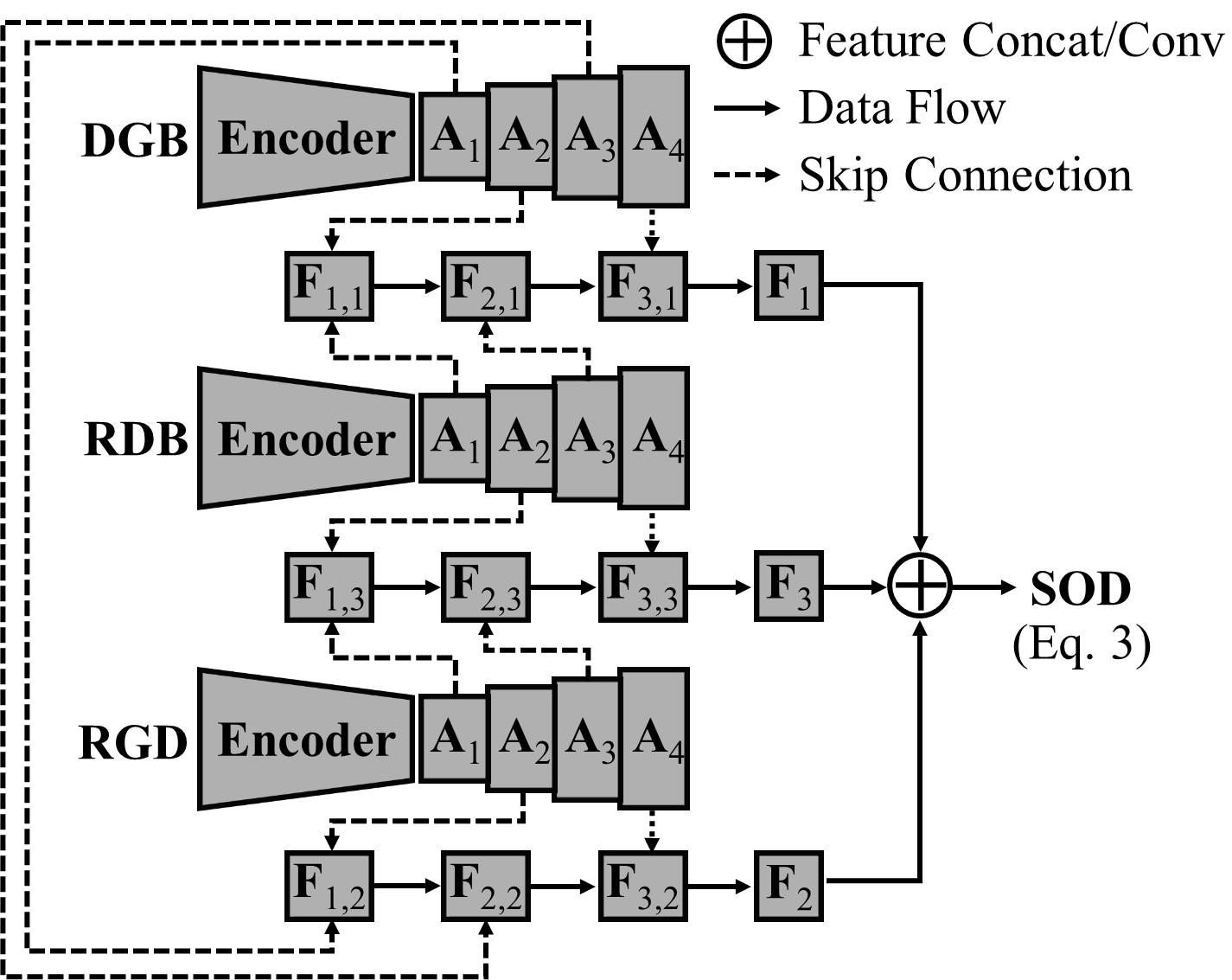}
  \caption{The detailed network architecture of our triple-stream fusion network, in which the multi-level features ($\textbf{A}_i, i\in\{1,2,3,4\}$) are directly obtained from 3 parallel backbone branches, see details in Fig.~\ref{fig:backbone}.}
  \label{Network}
\end{figure}

So far, we have obtained 3 independent branches by respectively feeding DBG, RDB, and RGD into 3 parallel backbone networks (Fig.~\ref{fig:backbone}).
To achieve a complementary fusion status among these 3 subnets, here, we propose a novel lightweight scheme to conduct multi-level and multi-net deep fusion, see the overall network architecture in Fig.~\ref{Network}.

Since the 3-dimensional input data for each subbranch only consists 2/3 color information (1/3 depth channel), we shall simultaneously resort multi-level deep features which are obtained from different branches to make full use of ``all'' color information.
For example, we concatenate the deep feature $\textbf{A}_1$ in the DGB branch with the $\textbf{A}_2$ in the RGD branch and then convolve it as the fused $\textbf{F}_{1,2}$.
Then, we sequentially combine $\textbf{F}_{1,2}$ with the deep feature $\textbf{A}_3$ in the DGB branch as the $\textbf{F}_{2,2}$, and the $\textbf{F}_{2,2}$ is further combined with the $\textbf{A}_4$ in the RGD branch as the $\textbf{F}_{3,2}$, which is finally convoluted as the fused saliency $\textbf{F}_2$.

Once all these branch-wise saliency estimations have been obtained, we directly convolve $\textbf{F}_1$, $\textbf{F}_2$, and $\textbf{F}_3$ as the final detection result (see Eq.~\ref{eq:Sal1}).
\begin{equation}
\label{eq:Sal1}
\mathrm{SOD}=f(\textbf{F}_1\circ \textbf{F}_2 \circ \textbf{F}_3).
\end{equation}

For each fusion procedure, we have applied one loss function over it to ensure the correctness of the computed deep features.
Thus, there are totally \{1 (SOD) + 3 (backbones) * 4 (levels) + 3 (branches) * 3 (levels) = 22\} loss functions in our triple-stream network, in which the total loss function can be formulated as Eq.~\ref{eq:Loss}.
\begin{equation}
\label{eq:Loss}
\begin{aligned}
\mathrm{Loss}=\alpha_{0}\cdot{\rm L}({\rm SOD})&+\sum_{j=1}^{3} \sum_{i=1}^{4}\alpha_{i,j}\cdot{\rm L}(\textbf{A}_{i,j}) \\&+ \sum_{r=1}^{3} \sum_{k=1}^{3}\alpha_{k,r}\cdot{\rm L}(\textbf{F}_{k,r}).
\end{aligned}
\end{equation}

Here $\rm Loss$ denotes the total loss of our network, and $\rm L(*)$ denotes the cross-entropy loss function;
$\alpha_*$ denote the weights of different loss functions, in which we empirically set $\alpha_{1,j}$, $\alpha_{2,j}$, $\alpha_{3,j}$, $\alpha_{4,j}$ and $\alpha_0$ to \{0.6, 0.7, 0.8, 0.9, 1\} respectively and $\alpha_{*,r}$ are set to 0.9.

Specifically, each of these loss functions is proved to be indispensable in Sec.~\ref{Sec:Multi-Loss} and Table~\ref{tab:loss}.

\subsection{Network with Saliency Recurrent}
\label{sec:re}
The primary motivation of our method is to use the depth channel to complement RGB information for a high-performance salient object detection result.
Our triple-stream network has already produced high-quality RGB-D saliency maps, which can be regarded as ``corrected'' depth maps to boost the overall SOD performance.
So we introduce a recurrent stage to \emph{\textbf{refine}} our saliency maps.
We use these saliency maps to replace the aforementioned depth part; i.e., we convert the original \{\textbf{D}GB, R\textbf{D}B, RG\textbf{D}\} into \{\textbf{S}GB, R\textbf{S}B, RG\textbf{S}\}, where \textbf{S} denotes the saliency prediction made by our triple-stream network (SOD, Eq.\ref{eq:Sal1}).
Then, we recursively feed \{SGB, RSB, RGS\} into our triple-stream network again to further improve the detection performance ($\rm SOD^+$, Eq.\ref{eq:Sal2}), and the quantitative results showing its effectiveness can be found in Table~\ref{table:component}.
\begin{equation}
\label{eq:Sal2}
\mathrm{SOD}^+={\rm TriNet}\big(\rm R,G,B,{\rm TriNet}(R,G,B,D)\big),
\end{equation}
where, ${\rm TriNet}$ denotes our novel triple-stream network.

\section{Experiments}

\subsection{Implementation Details}
We leverage the Caffe toolbox to implement our method.
We train and test our model on a desktop computer with NVIDIA GTX 1080 GPU (8G memory), Intel Core I7-6700 CPU (4 cores with 8 threads, 3.40GHz) and 32G RAM.

For a fair comparison, we follow an identical training and testing protocol adopted in~\cite{zhao2019contrast}; i.e., our training set consists of 1400 stereo images from NJUDS~\cite{NJUDS} dataset and 650 stereo images from NLPR~\cite{ECCV_P2014} dataset, and the rest data is used as the testing dataset.

Our training includes three steps:
\underline{1)} We firstly train our 3 independent backbone nets by solely using the RGB component; \underline{2)} We fine-tune each sub-branch using the newly formulated input data, i.e., DGB, RDB, and RGD; \underline{3)} We jointly fine-tune the triple-stream fusion network.
For each step in the training phase, it takes 5000 iterations with SGD backpropagation.
The learning rate, weight decay, momentum and iter size are assigned to \{1e-7, 0.0005, 0.9, 10\} respectively.

In the testing phase, our model takes almost $0.064s$ (15.6 FPS) to conduct salient object detection for a single $224\times 224$ RGB-D image.

\vspace{-0.2cm}
\subsection{Dataset and Evaluation Metrics}
We have quantitatively evaluated our method over 5 widely used benchmark datasets, e.g., NJUDS~\cite{NJUDS}, NLPR~\cite{ECCV_P2014}, SSB~\cite{SSB}, SSD~\cite{SSD} and DES~\cite{DES}.
A brief introduction will be given below:

\textbf{NJUDS} \cite{NJUDS} contains 1985 pairs of color and depth images with manually labeled ground truth. Its color images are captured from the indoor environment, outdoor environment and stereo movies. The depth maps are generated by an optical flow method and we normalized them to [0,255].

\textbf{NLPR} \cite{ECCV_P2014} is captured by the Microsoft Kinect in both indoor and outdoor scenarios. It contains 1000 natural image pairs. Its depth maps are of high-quality and sharp boundaries to distinguish the foreground and background. We normalized the depth images to [0,255]. Meanwhile, we reversed the value of depth data so that the salient areas have a high depth value.

\textbf{SSB} \cite{SSB} is also named \textbf{STEREO} which contains 1000 RGB and depth image pairs. These pictures are captured from indoor places, outdoor nature scenes and stereo movies. However, lots of pictures in SSB are repeated in NJUDS and NLPR datasets.

\textbf{SSD} \cite{SSD} is selected from the stereo movies and consists of 80 indoor/outdoor stereo images. Most scenes in this dataset contain multiple salient objects.
Specifically, the human-annotated ground truths in this dataset is quite different from the other datasets.
In this dataset, scenes clearly contain multiple salient objects, while the saliency ground truth annotations of these images tend to regard only one of them as the salient one. The details are shown in Fig.~\ref{fig:Demo_SSD}.
Thus, this dataset is more challenging than others.

\textbf{DES} \cite{DES} is captured by the Microsoft Kinect and contains 135 indoor stereo images. However, due to the limited quality, the depth channel of most RGB-D images in this dataset may only be able to coarsely locate the salient object.


\textbf{Evaluation Metrics}
We have evaluated the performance of our method and other SOTA methods using 8 widely adopted metrics~\cite{fan2018SOC}, e.g., S-measure~\cite{fan2017structure}, E-measure~\cite{Fan2018Enhanced} (adaptive, mean, max), F-measure (adaptive, mean, max) and mean absolute error (MAE).

\begin{table}[t]
  \centering
  \caption{Bi-stream fusion biasing quantitative results, in which the ``SIN'',``BI'' and ``TRI'' are the abbreviations of ``single-stream'', ``bi-stream'' and ``triple-stream'' respectively.
  In general, both the traditional bi-stream fusion and the single/multiple stream SOTA methods will bias to the color information regardless of the quality of depth maps.
  In sharp contrast, our method can handle this problem well.
  All deep models mentioned are trained over an \textbf{identical} training set containing both high-quality and low-quality cases. We divide the testing set into the ``high-quality'' subset and the ``low-quality'' subset to highlight the biased fusion status of the conventional bi-stream networks.}
  \resizebox{1\columnwidth}{!}{
    \begin{tabular}{|c|r|cc|cc|}
    \hline
    \multicolumn{2}{|c|}{\multirow{2}[4]{*}{NLPR~\cite{ECCV_P2014} (300)}} & \multicolumn{2}{c|}{Low-quality Depth: 246} & \multicolumn{2}{c|}{High-quality Depth: 54} \bigstrut\\
\cline{3-6}    \multicolumn{2}{|c|}{} & \multicolumn{1}{c|}{D-Branch} & C-Branch & \multicolumn{1}{c|}{D-Branch} & C-Branch \bigstrut\\
    \hline
    \multirow{8}[10]{*}{\begin{sideways}\textbf{MAE}\end{sideways}} & Before Fusion & \multicolumn{1}{c|}{.067} & .032  & \multicolumn{1}{c|}{\textbf{.058}} & .096 \bigstrut\\
\cline{2-6}          & After Fusion & \multicolumn{2}{c|}{.034} & \multicolumn{2}{c|}{.085} \bigstrut\\
\cline{2-6}          & Ours/TRI~\textcolor[rgb]{ 1,  1,  1}{[22]} & \multicolumn{2}{c|}{.027} & \multicolumn{2}{c|}{\textbf{.056}} \bigstrut[t]\\
          & TANet/TRI~\cite{chen2019three} & \multicolumn{2}{c|}{.033} & \multicolumn{2}{c|}{.073} \bigstrut[b]\\
\cline{2-6}          & PCA/BI~\cite{chen2018progressively} & \multicolumn{2}{c|}{.036} & \multicolumn{2}{c|}{.080} \bigstrut[t]\\
          & MMCI/BI~\cite{chen2019multi} & \multicolumn{2}{c|}{.053} & \multicolumn{2}{c|}{.088} \\
          & AFNet/BI~\cite{wang2019adaptive} & \multicolumn{2}{c|}{.052} & \multicolumn{2}{c|}{.089} \bigstrut[b]\\
\cline{2-6}          & CPFP/SIN~\cite{zhao2019contrast} & \multicolumn{2}{c|}{.028} & \multicolumn{2}{c|}{.068} \bigstrut\\
    \hline
    \multicolumn{2}{|c|}{\multirow{2}[4]{*}{DES~\cite{DES} (135)}} & \multicolumn{2}{c|}{Low-quality Depth: 83} & \multicolumn{2}{c|}{High-quality Depth: 52} \bigstrut\\
\cline{3-6}    \multicolumn{2}{|c|}{} & \multicolumn{1}{c|}{D-Branch} & C-Branch & \multicolumn{1}{c|}{D-Branch} & C-Branch \bigstrut\\
    \hline
    \multirow{8}[10]{*}{\begin{sideways}\textbf{MAE}\end{sideways}} & Before Fusion & \multicolumn{1}{c|}{.042} & .024  & \multicolumn{1}{c|}{\textbf{.044}} & .084 \bigstrut\\
\cline{2-6}          & After Fusion & \multicolumn{2}{c|}{.025} & \multicolumn{2}{c|}{.072} \bigstrut\\
\cline{2-6}          & Ours/TRI~\textcolor[rgb]{ 1,  1,  1}{[22]} & \multicolumn{2}{c|}{.018} & \multicolumn{2}{c|}{\textbf{.049}} \bigstrut[t]\\
          & TANet/TRI~\cite{chen2019three} & \multicolumn{2}{c|}{.030} & \multicolumn{2}{c|}{.071} \bigstrut[b]\\
\cline{2-6}          & PCA/BI~\cite{chen2018progressively} & \multicolumn{2}{c|}{.030} & \multicolumn{2}{c|}{.079} \bigstrut[t]\\
          & MMCI/BI~\cite{chen2019multi} & \multicolumn{2}{c|}{.050} & \multicolumn{2}{c|}{.089} \\
          & AFNet/BI~\cite{wang2019adaptive} & \multicolumn{2}{c|}{.053} & \multicolumn{2}{c|}{.092} \bigstrut[b]\\
\cline{2-6}          & CPFP/SIN~\cite{zhao2019contrast} & \multicolumn{2}{c|}{.021} & \multicolumn{2}{c|}{.064} \bigstrut\\
    \hline
    \end{tabular}%
    }
  \label{tab:BiasingProof}%
  \vspace{-0.2cm}
\end{table}%

\begin{table}[t]
\caption{Component evaluation results over NLPR dataset.
``RGB'', ``D'', ``DGB'', ``RDB'' and ``RGD'' denotes different input data combination of single/bi-stream/triple-stream network.
``LC'' and ``MF'' represent the different fusion strategies, e.g., linear concatenation and the proposed multi-level and multi-net deep feature fusion model.
$\uparrow$ means that the larger one is better, and $\downarrow$ denotes that the smaller one is better.}
  \centering
  \resizebox{1\columnwidth}{!}{
    \begin{tabular}{|cc|ccc|}
    \hline
    \multicolumn{2}{|c|}{\textbf{\textbf{\ \ \ \ Combinations}}}  & meanF $\uparrow$ & maxF $\uparrow$ & MAE $\downarrow$ \\
    \hline
    \multirow{6}[2]{*}{\begin{sideways}Single-stream\end{sideways}} & \{RGB\} & .805 & .837 & .051 \bigstrut[t]\\
          & \{D\}   & .712 & .743 & .115 \\
          & \{RGBD\} & .772 & .826 & .053 \\
          & \{DGB\} & .821 & .853 & .045 \\
          & \{RDB\} & .819 & .853 & .043 \\
          & \{RGD\} & .821 & .852 & .042 \bigstrut[b]\\
    \hline
    \multirow{4}[2]{*}{\begin{sideways}Bi-Stream\end{sideways}} & \{RGB+D\} & .825 & .853 & .042 \bigstrut[t]\\
          & \{DGB+RDB\} & .827 & .858 & .040 \\
          & \{RDB+RGD\} & .831 & .862 & .039 \\
          & \{RGD+DGB\} & .832 & .861 & .039 \bigstrut[b]\\
    \hline
    \multirow{10}[4]{*}{\begin{sideways}\ \ Tri-Stream\end{sideways}} & LC\{RGB+D+D\} & .822 & .851 & .041 \bigstrut[t]\\
          & LC\{RGB+RGB+D\} & .829 & .860  & .040 \\
          & LC\{DGB+RDB+RGD\} & .838 & .862 & .038 \\
          & MF\{RGB+D+D\} & .828 & .858 & .040 \\
          & MF\{GB+RB+RG\} & .832 & .861 & .038 \\
          & MF\{RGB+RGB+D\} & .834 & .865 & .038 \\
          & MF\{RGB+RGB+RGB\} & .835 & .865 & .037 \\
          & MF\{DGB+RDB+RGD\} & .844 & .872 & .035 \bigstrut[b]\\
         & \textbf{Final} (Recurrent) & .854 & .880 & .032 \bigstrut\\
    \hline
    \end{tabular}%
    }
\vspace{-0.2cm}
  \label{table:component}%
\end{table}%


\subsection{Ablation Experiments}
In order to validate the effectiveness of our method, we have conducted the ablation experiments via F-measure and MAE metrics regarding ``Bi-Stream Fusion Biasing Quantitative Results'' (Table~\ref{tab:BiasingProof}), ``Effectiveness of Data-level Fusion'' (Table~\ref{table:component}), ``Reasoning behind Channel-Wise Data Fusion'' (Table~\ref{tab:Space}), ``Effectiveness of the Proposed Triple-Stream Fusion Strategy'' (Table~\ref{table:component}), ``Lightweight Fusion Strategy'' (Table~\ref{tab:modelsize}~\ref{tab:connection}) and ``Effectiveness of Multiple Loss Functions'' (Table~\ref{tab:loss}).

\textbf{Bi-Stream Fusion Biasing Quantitative Results}.
\label{Sec:BiasingProof}
The biasing tendency in the conventional bi-stream fusion networks (see the first two rows in Table~\ref{tab:BiasingProof}) is mainly induced by the unbalanced status between their two sub-branches.
As shown in the ``high-quality depth'' columns, we can see that there are total 300 images in the NLPR testing set, and only 54 depth maps are capable of producing more distinguish salient features than RGB images. The strong data adaptability of the current deep learning based techniques leads the bi-stream fusion subnet to bias towards the informative RGB component (MAE is \underline{0.096}), producing a mediocre detection (MAE is 0.085) even in the case that the depth channel is trustworthy (MAE is \underline{0.058}).
On the other hand, we can see from the 4th to 7th rows that most of the SOTA methods (including bi-stream PCA~\cite{chen2018progressively}, MMCI~\cite{chen2019multi}, AFNet~\cite{wang2019adaptive} and triple-stream TANet~\cite{chen2019three}) bias towards the color information even if the depth information is more reliable in some cases.
Benefited from the single-stream structure, we can see that CPFP~\cite{zhao2019contrast} (the eighth row in Table~\ref{tab:BiasingProof}) decreases the tendency of biasing but it doesn't solve the problem.
Compared with the abovementioned SOTA methods, our model overcomes this weakness by balancing its sub-branches via balanced input data, achieving a much improved performance in cases with either high-quality (up to \underline{37}\% improvement in maxF) or low-quality (up to \underline{47}\% improvement in maxF) depth information (see the third row in Table~\ref{tab:BiasingProof}).

\textbf{Effectiveness of Data-level Fusion}.
We design traditional single/bi-stream/triple-stream networks based on VGG16 network, the fusion strategies of the bi-stream/tri-stream networks are either linear concatenation (denoted by LC) or the proposed multi-level and multi-net deep feature fusion model (denoted by MF).
To validate the effectiveness of the proposed data-level fusion scheme, we feed different input data into the single-stream network respectively.
As shown in Table~\ref{table:component}, our data-level fusion scheme, e.g., DGB, RDB and RGD, achieves obvious performance improvements than the original RGB saliency maps (\underline{1.9}\% in maxF) and depth saliency maps (\underline{14.8}\% in maxF).
Also, we may easily notice that the single-stream network using our novel data-level fusion scheme has achieved comparable performance to the bi-stream network using the conventional RGB-D input data.
Meanwhile, as shown in the middle rows of Table~\ref{table:component}, the conventional bi-stream network using our data-level fused input can get significant performance improvements, e.g., the \{RDB$+$RGD\} improves the conventional \{RGB$+$D\} almost \underline{1}\% in maxF.
Actually, we believe that the above results are mainly induced by the following three aspects:\\
\underline{1)} The conventional bi-stream network, which takes depth channel as the sole input for its depth branch, can not make full use of its backbone network, failing to obtain deep features with meaningful semantic information; \\
\underline{2)} The conventional bi-stream network easily biases its fusion procedure to the color branch due to the worse performance of its depth branch;\\
\underline{3)} Our data-level fusion is simple yet effective, which is able to complement the color information with the depth information. Meanwhile, it can well adapt to the backbone network because of the 3-dimensional data structure, obtaining useful semantic information for saliency detection.

On the other hand, each of the 3-dimensional data outperforms the 4-dimensional RGBD data in the single-stream network, even though the 4-dimensional data includes one more channel information.
The main reason is that the widely used feature backbones are pre-trained using the 3-dimensional training set with strong semantic information, while such pre-learned semantic information may get lost if we fine-tune it over 4-dimensional data (i.e., RGBD), showing another advantage of our data-wise fusion.

\textbf{Reasoning behind Channel-Wise Data Fusion}.
As we all know, the performance of feature backbones usually plays an important role in determining the overall detection performance, and the feature backbones used in the salient object detection field are usually trained over other large-scale training sets (e.g., the ImageNet).
Since the function of depth channel is quite similar to RGB channels, providing some possible cues to separate different objects, and thus the feature gap between the original RGB and the newly re-formulated data (\{D+RG\}, \{D+RB\}, and \{D+GB\}) shall be marginal.
So, by using such re-formulated data, the pre-trained feature-backbones are still capable of providing meaningful and discriminative deep features even after being fine-tuned using other training sets, e.g., the widely-used RGB-D training set with 2050 images.
However, there usually exist large differences between the 3-dimensional RGB and the 2-dimensional R+D (or G/B+D), which make the pre-trained feature backbones not suitable for the 2-dimensional input data, requiring a complete new training.
Unfortunately, the widely-used RGB-D training set only consists of 2050 training instances, which are clearly insufficient to train a complete new feature backbone.
Thus, due to the abovementioned aspects, we decide to resort the 3-dimensional data formulations, i.e., \{D+RG\}, \{D+RB\}, and \{D+GB\}.

Specifically, because the off-the-shelf feature backbones are all trained using RGB training set, it will lead to significant performance degeneration if we choose to use the luminance \& chrominance formulation (i.e., the YUV color space).
To make the above explanations more convincing, we have tested the performance of the above mentioned 2-dimensional case and the YUV case.
As shown in Table~\ref{tab:Space}, the second row has achieved the best performance, while other cases can not perform well because of the large gap between these data formulations and the original RGB training space.

We also verify the advantage of using depth information via quantitative comparisons, i.e., the proposed network \emph{without using} depth information (denoted as MF\{GB+RB+RG\} and MF\{RGB+RGB+RGB\}) vs. the proposed network \emph{using} depth information (denoted as MF\{DGB+RDB+RGD\}) .
As shown in Table~\ref{table:component}, the combination using depth information has achieved the best performance, while the other two combinations without using depth information have exhibited worse performances.
Since all these results are obtained via an identical network structure, the performance margins are mainly induced by the depth information, showing the advantages of using depth information.

\begin{table}[t]
  \centering
  \caption{Ablation experiments of different data formulations. We highlight the best results with bold typeface.}
  \resizebox{1\columnwidth}{!}{
    \begin{tabular}{|c|ccc|ccc|ccc|}
    \hline
    Color Space & \multicolumn{3}{c|}{\textbf{NJUDS}} & \multicolumn{3}{c|}{\textbf{NLPR}} & \multicolumn{3}{c|}{\textbf{DES}} \bigstrut\\
\cline{2-10}    RGB/YUV & meanF & maxF & MAE & meanF & maxF & MAE & meanF & maxF & MAE \bigstrut\\
\hline
    RD+GD+BD & .841  & .875  & .062  & .840  & .870  & .038  & .817  & .848  & .036 \bigstrut[t]\\
    RGD+RDB+DGB & \textbf{.853} & \textbf{.881} & \textbf{.058} & \textbf{.844} & \textbf{.872} & \textbf{.035} & \textbf{.846} & \textbf{.869} & \textbf{.031} \\
    YUD+YDV+DUV & .741  & .756  & .100  & .740  & .767  & .055  & .646  & .775  & .063 \bigstrut[b]\\
    \hline
    \end{tabular}%
  \label{tab:Space}%
  }
\end{table}%

\begin{table}[t]
  \centering
  \caption{
  Comparisons regarding the number of parameters between our model and other SOTA methods (unit: million).
  From the perspective of fusion-total ratio, our fusion scheme achieves the strong inter-subnet interactions yet at the lowest computational cost.}
    \resizebox{1\columnwidth}{!}{
    \begin{tabular}{|r|c|c|c|c|c|}
    \hline
    Method & Total & Backbone & Fusion & F-T Ratio & FPS \bigstrut\\
    \hline
    CPFP~\cite{zhao2019contrast} & 72.90 & 1 $\times$ 15 & 57.90 & 79\%  & 9.2 \bigstrut[t]\\
    TESF~\cite{chen2020improved} & 98.30 & 1 $\times$ 15 & 83.30 & 85\%  & 9.8 \\
    PCA~\cite{chen2018progressively} & 138.7 & 2 $\times$ 15 & 108.7 & 78\%  & 15.3 \\
    DCFF~\cite{chen2019discriminative} & 244.8 & 2 $\times$ 15 & 214.8 & 88\%  & 13.1 \\
    MMCI~\cite{chen2019multi} & 241.7 & 2 $\times$ 15 & 211.7 & 88\%  & 19.5 \\
    TANet~\cite{chen2019three} & 247.5 & 3 $\times$ 15 & 202.5 & 82\%  & 14.2 \\
    Our Model & 91.80 & 3 $\times$ 15 & 46.80 & 51\%  & 15.6 \bigstrut[b]\\
    \hline
    \end{tabular}%
    }
  \label{tab:modelsize}%
  \vspace{-0.2cm}
\end{table}%

\begin{figure*}[t]
\centering
\includegraphics[width=1\linewidth]{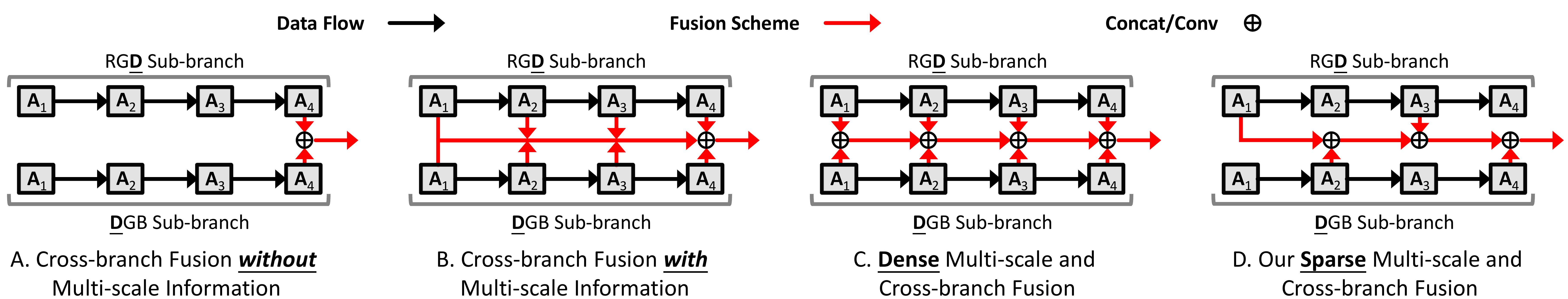}
\caption{Comparisons between different fusion schemes, in which $\textbf{A}_i, i\in\{1,2,3,4\}$ represent multi-level features of our backbones, which are parallel identical to that in Fig.~\ref{Network}.
Our fusion scheme is shown as subfigure D, which is lightweight designed and well-performed, please refer to the corresponding quantitative results in Table~\ref{tab:connection}.}
\label{fig:FusionScheme}
\end{figure*}

\textbf{Effectiveness of the Proposed Triple-Stream Fusion Strategy} (Sec.~\ref{Sec:fusion}).
We also explore the effectiveness of the proposed multi-level and multi-net deep feature fusion strategy.
As shown in Table~\ref{table:component}. Tri-Stream, the proposed fusion strategy achieves a significant performance improvement; i.e., the proposed fusion strategy (denoted by \underline{MF}) outperforms the conventional linear concatenation (denoted by \underline{LC}) using an identical input data (\underline{0.8}\%/\underline{0.6}\%/\underline{1.2}\% in maxF over \{RGB+D+D\}, \{RGB+RGB+D\} and \{DGB+RDB+RGD\} respectively).
Compared to the conventional fusion scheme, the proposed multi-level and multi-net feature fusion scheme has two major advantages:\\
\underline{1)} It achieves an optimal complementary fusion status among its triple-stream sub-branches;\\
\underline{2)} It fuses the multi-scale deep features of different sub-branches, and thus its final prediction can well reserve tiny details.

\begin{table}[t]
  \centering
  \caption{Ablation study over triple-stream fusion schemes.
  The detailed structures of different fusion schemes are shown in Fig.~\ref{fig:FusionScheme}.
  The unit of the number of parameters is ``million''.
  We highlight the best performances with bold typeface.}
   \resizebox{1\columnwidth}{!}{
	\begin{tabular}{|c|c|ccc|ccc|ccc|}
  \hline
	\multirow{2}[4]{*}{Fig.~\ref{fig:FusionScheme}} & \textbf{Para-} & \multicolumn{3}{c|}{\textbf{NJUDS}} & \multicolumn{3}{c|}{\textbf{NLPR}} & \multicolumn{3}{c|}{\textbf{DES}} \bigstrut\\
	\cline{3-11}      & \textbf{meters} & meanF & maxF & MAE & meanF & maxF & MAE  & meanF & maxF & MAE \bigstrut\\
  \hline
	A     & ~86.2  & .841 & .865 & .063 & .838 & .863 & .038 & .836 & .859 & .039 \bigstrut[t]\\
	B     & ~90.4  & .849 & .876 & .061 & .841 & .868 & .037 & .840 & .864 & .034 \\
	C     & 144.9 & \textbf{.855} & \textbf{.882} & \textbf{.057} & \textbf{.845} & .871 & \textbf{.035} & \textbf{.846} & .868 & \textbf{.031} \\
	D     & ~91.8  & .853 & .881 & .058 & .844 & \textbf{.872} & \textbf{.035} & \textbf{.846} & \textbf{.869} & \textbf{.031} \bigstrut[b] \\
  \hline
	\end{tabular}
	}
  \label{tab:connection}%
\end{table}%

\begin{table}[t]
  \centering
  \caption{Ablation study over different combinations of loss functions.
  We divide the loss functions into three parts (Loss A, F, S) according to their applied stages.
  The loss weights are assigned as Eq.~\ref{eq:Loss}.
  $\rm{L_{S+F+A}^*}$ denotes that we set all hyperparameters in the total loss into 1, equally.
  The best performances are highlighted with bold typeface.}
  \resizebox{1\columnwidth}{!}{
  \begin{tabular}{|l|ccc|ccc|ccc|}
    \hline
    \multicolumn{1}{|c|}{Loss} & \multicolumn{3}{c|}{\textbf{NJUDS}} & \multicolumn{3}{c|}{\textbf{NLPR}} & \multicolumn{3}{c|}{\textbf{DES}} \bigstrut\\
\cline{2-10}    \multicolumn{1}{|c|}{Functions} & meanF & maxF & MAE & meanF & maxF & MAE & meanF & maxF & MAE \bigstrut\\
    \hline
    $\rm{L_S}$    & .813 & .842 & .077 & .79  & .821 & .049 & .767 & .826 & .049 \bigstrut[t]\\
    $\rm{L_{S+A}}$  & .852 & .875 & .061 & .832 & .858 & .038 & .815 & .842 & .036 \\
    $\rm{L_{S+F}}$  & .848 & .867 & .062 & .818 & .842 & .040 & .790 & .828 & .039 \\
    $\rm{L_{S+F+A}}$ & \textbf{.853} & \textbf{.881} & \textbf{.058} & \textbf{.844} & \textbf{.872} & \textbf{.035} & \textbf{.846} & \textbf{.869} & \textbf{.031} \\
    $\rm{L_{S+F+A}^*}$ & .843 & .868 & .062 & .810 & .835 & .041 & .788 & .823 & .040 \bigstrut[b]\\
    \hline
    \end{tabular}%
    }
  \label{tab:loss}%
  \vspace{-0.2cm}
\end{table}%

\textbf{Lightweight Fusion Strategy}.
To validate the lightweight structure of the proposed fusion strategy, we \underline{1)} compare the number of parameters with other SOTA methods and \underline{2)} compare the performance over evaluation metrics with dense fusion schemes.
As shown in Table~\ref{tab:modelsize}, we have listed the parameter quantity in the backbone and fusion subnetwork individually to show the computational expenses. All the compared methods are based on the VGG network.
The unit of the number of parameters is \emph{``million''}.
At the first glance, the proposed method has a larger number of parameters than single-stream CPFP~\cite{zhao2019contrast}, i.e., the number of parameters in CPFP is about 15 (One Backbone) $+$ 57.9 (Fusion) $=$ \underline{72.9} in total, yet the number of parameters in our model is 3 $\times$ 15 (Three Backbones) $+$ 46.8 (Fusion) $=$ \underline{91.8} in total.
Since our method has adopted the triple-stream network architecture which consists of 3 backbones, and these backbones make up almost \underline{49}\% parameters and the fusion connections make up the rest \underline{51}\%, which is a much lower ratio than the CPFP method (\underline{79}\%).
Thus, from the perspective of a triple-stream network, our model is lightweight designed and the fusion connections of our method are very sparse (with only 12 short-connections shown in Fig.~\ref{Network}).
Our model achieves a speed of 15.6 FPS, which is faster than most of the STOA methods.


Meanwhile, we compare our method with other triple-stream fusion schemes, which are shown in Fig.~\ref{fig:FusionScheme}.
Notice that we only take the fusion procedures between \textbf{D}GB and RG\textbf{D} branches for example.
Since the 3-dimensional input data for each subbranch only consists of 2 color channels and 1 depth channel, we sparsely combine \{DGB, RDB\}, \{RDB, RGD\} and \{RGD, DGB\} branches first, and then iteratively integrate the output of these three subbranches to obtain our final detection.

As shown in Fig.~\ref{fig:FusionScheme}A and B, we evaluate the performance of the fusion scheme without/with using the multi-scale information.
Compared with our fusion scheme, Fig.~\ref{fig:FusionScheme}C shows a more dense manner to fuse multi-scale and multi-net features.
The performances of different fusion schemes are respectively reported in Table~\ref{tab:connection}.
Compared the fusion manner C and D (i.e., the last two rows in Table~\ref{tab:connection}), it is obvious that it may achieve about \underline{0.4}\% performance improvement if we take all multi-scale and multi-net features into consideration.
We can observe that our sparse fusion scheme (Fig.~\ref{fig:FusionScheme}D) performs comparable to the dense fusion scheme (Fig.~\ref{fig:FusionScheme}C), saving almost 37\% parameters in total because our proposed fusion strategy considers all channel information and different scale features by a crossing connection manner.
So, as one of the key contributions, our fusion scheme achieves the strong inter-subnet interactions yet at the lowest computational cost, and this is why we claim our method as a lightweight design.

\textbf{Effectiveness of Multiple Loss Functions}.
\label{Sec:Multi-Loss}
There are 22 loss functions in our model, though a large number in quantity, all of them are indispensable.
The usage of such a large amount of loss functions is inspired by the classic DSS~\cite{CVPR_H2017}, in which the single-stream method has adopted 6 loss functions in total to ensure high-quality side-outputs.
The DSS has proved that discriminative features in different hidden layers will complement each other, and it is required to assign an individual loss function for each hidden layer to obtain such features.

Many existed models have followed the DSS, but they fail to explore the performance improvement by using multiple loss functions in different stages.
To clearly distinguish these loss functions (Eq.~\ref{eq:Loss}), we divide them into three parts according to their supervised stages; i.e., we call the loss functions using in the backbone/branch-fusion/final-result as ``$\rm{L_A}$ (i.e., $\rm L(\textbf{A})$ in Eq.~\ref{eq:Loss})'', ``$\rm{L_F}$ (i.e., $\rm L(\textbf{F})$ in Eq.~\ref{eq:Loss})'', ``$\rm{L_S}$ (i.e., $\rm L(SOD)$ in Eq.~\ref{eq:Loss})'' respectively.
Then we make an extensive ablation study to validate the effectiveness of these loss functions.

As shown in Table~\ref{tab:loss}, we regard the $\rm{L_A}$ as the baseline, and other loss functions will be applied to it respectively.
With the increasing of loss functions, the network makes a steady performance improvement (i.e., \underline{6.2}\% in maxF, which can be seen in the first four rows in Table~\ref{tab:loss}).
Compared with the loss functions in branch fusion stage (marked as $\rm{L_{S+F}}$), the loss functions in the backbone stage (marked as $\rm{L_{S+A}}$) have leaded a lager performance improvement (\underline{1.7}\% in maxF).

In order to validate the effectiveness of our empirical assignment for those hyper-parameters in Eq.~\ref{eq:Loss}, we have conducted another verification, in which we set all the hyper-parameters into 1 (marked as $\rm{L_{S+F+A}^*}$ in the 5th row of Table~\ref{tab:loss}).
Obviously, our method is slightly sensitive to these hyper-parameters, floating its overall performance within about \underline{4.4}\%.


\begin{table*}[t]
  \centering
  \caption{Quantitative comparison results in terms of S-measure, E-measure, MAE and F-measure over 5 challenging benchmark datasets. $\uparrow$ denotes larger is better, and $\downarrow$ denotes smaller is better. The best results are highlighted with bold typeface.
  The advantage of our method is not much obvious over the SSD dataset (still belonging to the top-3 methods), the reason is discussed in section~\ref{Sec:PerformanceComparisons}.}
  \resizebox{2\columnwidth}{!}{
    \begin{tabular}{l|l|cccccccccccccccc|cc}
    \toprule
    \multirow{2}[2]{*}{} & \multirow{2}[2]{*}{\textbf{Metric}} & {LHM} & {ACSD} & {GP} & {LBE} & {DCMC} & {SE} & {CDCP} & {MDSF} & {DF} & {CDB} & {CTMF} & {PCA} & {AFNet} & {MMCI} & {TANet} & {CPFP} & {Ours} & {Ours} \\
          &       & {2014} & {2014} & {2015} & {2016} & {2016} & {2016} & {2017} & {2017} & {2017} & {2018} & {2018} & {2018} & {2019} & {2019} & {2019} & {2019} & {SOD} & {$\rm SOD^+$} \\
    \midrule
    \multirow{8}[2]{*}{{\begin{sideways}\textbf{NJUDS~\cite{NJUDS}}\end{sideways}}} & {\textbf{Sm} $\uparrow$} & .514  & .699  & .527  & .695  & .686  & .664  & .669  & .748  & .763  & .624  & .849  & .877  & .772  & .858  & .878  & .878 & .886 & \textbf{.886} \\
          & {\textbf{adpE} $\uparrow$} & .708  & .786  & .716  & .791  & .791  & .772  & .747  & .812  & .835  & .745  & .864  & .896 & .846  & .878  & .893  & .895  & .899 & \textbf{.901} \\
          & {\textbf{meanE} $\uparrow$} & .447  & .593  & .466  & .655  & .619  & .624  & .706  & .677  & .696  & .565  & .846  & .895  & .826  & .851  & .895  & \textbf{.910} & .901 & .909 \\
          & {\textbf{maxE} $\uparrow$} & .724  & .803  & .703  & .803  & .799  & .813  & .741  & .838  & .864  & .742  & .913  & .924  & .853  & .915  & .925 & .923  & .926 & \textbf{.926} \\
          & {\textbf{adpF} $\uparrow$} & .638  & .696  & .655  & .740  & .717  & .734  & .624  & .757  & .784  & .648  & .788  & .844 & .768  & .812  & .844 & .837  & .843  & \textbf{.849} \\
          & {\textbf{meanF} $\uparrow$} & .328  & .512  & .357  & .606  & .556  & .583  & .595  & .628  & .650  & .482  & .779  & .840  & .764  & .793  & .841  & .850 & .853 & \textbf{.858} \\
          & {\textbf{maxF} $\uparrow$} & .632  & .711  & .647  & .748  & .715  & .748  & .621  & .775  & .804  & .648  & .845  & .872  & .775  & .852  & .874  & .877 & .881 & \textbf{.883} \\
          & {\textbf{MAE} $\downarrow$} & .205  & .202  & .211  & .153  & .172  & .169  & .180  & .157  & .141  & .203  & .085  & .059  & .100  & .079  & .060  & \textbf{.053} & .058 & .055 \\
    \midrule
    \multirow{8}[2]{*}{{\begin{sideways}\textbf{STERE~\cite{SSB}}\end{sideways}}} & {\textbf{Sm} $\uparrow$} & .562  & .692  & .588  & .660  & .731  & .708  & .713  & .728  & .757  & .615  & .848  & .875  & .825  & .873  & .871  & .879 & .883 & \textbf{.888} \\
          & {\textbf{adpE} $\uparrow$} & .770  & .793  & .784  & .749  & .831  & .825  & .796  & .830  & .838  & .808  & .864  & .897  & .886  & .901  & .906  & .903  & .911 & \textbf{.915} \\
          & {\textbf{meanE} $\uparrow$} & .484  & .592  & .509  & .601  & .655  & .665  & .751  & .614  & .691  & .561  & .841  & .887  & .872  & .873  & .893  & \textbf{.912} & .898 & .911 \\
          & {\textbf{maxE} $\uparrow$} & .771  & .806  & .743  & .787  & .819  & .846  & .786  & .809  & .847  & .823  & .912  & .925  & .887  & .927 & .923  & .925 & .924  & \textbf{.929} \\
          & {\textbf{adpF} $\uparrow$} & .703  & .661  & .711  & .595  & .742  & .748  & .666  & .744  & .742  & .713  & .771  & .826  & .807  & .829  & .835 & .830  & .837 & \textbf{.845} \\
          & {\textbf{meanF} $\uparrow$} & .378  & .478  & .405  & .501  & .590  & .610  & .638  & .527  & .617  & .489  & .758  & .818  & .806  & .813  & .828  & .841 & .838 & \textbf{.850} \\
          & {\textbf{maxF} $\uparrow$} & .683  & .669  & .671  & .633  & .740  & .755  & .664  & .719  & .757  & .717  & .831  & .860  & .823  & .863  & .861  & .874 & .871 & \textbf{.878} \\
          & {\textbf{MAE} $\downarrow$} & .172  & .200  & .182  & .250  & .148  & .143  & .149  & .176  & .141  & .166  & .086  & .064  & .075  & .068  & .060  & .051 & .055 & \textbf{.050} \\
    \midrule
    \multirow{8}[2]{*}{{\begin{sideways}\textbf{DES~\cite{DES}}\end{sideways}}} & {\textbf{Sm} $\uparrow$} & .578  & .728  & .636  & .703  & .707  & .741  & .709  & .741  & .752  & .645  & .863  & .842  & .770  & .848  & .858  & .872 & \textbf{.896} & .895 \\
          & {\textbf{adpE} $\uparrow$} & .761  & .855  & .785  & .911  & .849  & .852  & .816  & .869  & .877  & .868  & .911  & .912  & .874  & .904  & .919  & .927 & \textbf{.958} & .954 \\
          & {\textbf{meanE} $\uparrow$} & .477  & .612  & .503  & .649  & .632  & .707  & .748  & .621  & .684  & .572  & .826  & .838  & .810  & .825  & .863  & .889 & .902 & \textbf{.910} \\
          & {\textbf{maxE} $\uparrow$} & .653  & .850  & .670  & .890  & .773  & .856  & .811  & .851  & .870  & .830  & .932 & .893  & .881  & .928  & .910  & .923  & \textbf{.947} & .940 \\
          & {\textbf{adpF} $\uparrow$} & .631  & .717  & .686  & .796  & .702  & .726  & .625  & .744  & .753  & .729  & .778  & .782  & .730  & .762  & .795  & .829 & \textbf{.874} & .868 \\
          & {\textbf{meanF} $\uparrow$} & .345  & .513  & .412  & .576  & .542  & .617  & .585  & .523  & .604  & .502  & .756  & .765  & .713  & .735  & .790  & .824 & .846 & \textbf{.852} \\
          & {\textbf{maxF} $\uparrow$} & .511  & .756  & .597  & .788  & .666  & .741  & .631  & .746  & .766  & .723  & .844  & .804  & .729  & .822  & .827  & .846 & .869 & \textbf{.869} \\
          & {\textbf{MAE} $\downarrow$} & .114  & .169  & .168  & .208  & .111  & .090  & .115  & .122  & .093  & .100  & .055  & .049  & .068  & .065  & .046  & .038 & .031 & \textbf{.030} \\
    \midrule
    \multirow{8}[2]{*}{{\begin{sideways}\textbf{NLPR~\cite{ECCV_P2014}}\end{sideways}}} & {\textbf{Sm} $\uparrow$} & .630  & .673  & .654  & .762  & .724  & .756  & .727  & .805  & .802  & .629  & .860  & .874  & .799  & .856  & .886  & .888 & .899 & \textbf{.903} \\
          & {\textbf{adpE} $\uparrow$} & .813  & .742  & .804  & .855  & .786  & .839  & .800  & .812  & .868  & .809  & .869  & .916  & .884  & .872  & .916  & .924 & .934 & \textbf{.936} \\
          & {\textbf{meanE} $\uparrow$} & .560  & .578  & .571  & .719  & .684  & .742  & .781  & .745  & .755  & .565  & .840  & .887  & .851  & .841  & .902  & .918 & .913 & \textbf{.922} \\
          & {\textbf{maxE} $\uparrow$} & .766  & .780  & .723  & .855  & .793  & .847  & .820  & .885  & .880  & .791  & .929  & .925  & .879  & .913  & \textbf{.941} & .932  & .937 & .939 \\
          & {\textbf{adpF} $\uparrow$} & .664  & .535  & .659  & .736  & .614  & .692  & .608  & .665  & .744  & .613  & .724  & .795  & .747  & .730  & .796  & .823 & .839 & \textbf{.843} \\
          & {\textbf{meanF} $\uparrow$} & .427  & .429  & .451  & .736  & .543  & .624  & .609  & .649  & .664  & .422  & .740  & .802  & .755  & .737  & .819  & .840 & .844 & \textbf{.854} \\
          & {\textbf{maxF} $\uparrow$} & .622  & .607  & .611  & .745  & .648  & .713  & .645  & .793  & .778  & .618  & .825  & .841  & .771  & .815  & .863  & .867 & .872 & \textbf{.880} \\
          & {\textbf{MAE} $\downarrow$} & .108  & .179  & .146  & .081  & .117  & .091  & .112  & .095  & .085  & .114  & .056  & .044  & .058  & .059  & .041  & .036 & .035 & \textbf{.032} \\
    \midrule
    \multirow{8}[2]{*}{{\begin{sideways}\textbf{SSD~\cite{SSD}}\end{sideways}}} & {\textbf{Sm} $\uparrow$} & .566  & .675  & .615  & .621  & .704  & .675  & .603  & .673  & .747  & .562  & .776  & \textbf{.841} & .714  & .813  & .839 & .807  & .836 & .835 \\
          & {\textbf{adpE} $\uparrow$} & .730  & .765  & .795  & .729  & .786  & .778  & .705  & .772  & .812  & .737  & .838  & \textbf{.886} & .803  & .860  & .879 & .832  & .878  & .880 \\
          & {\textbf{meanE} $\uparrow$} & .498  & .566  & .529  & .574  & .646  & .631  & .676  & .576  & .690  & .477  & .796  & .856 & .762  & .796  & \textbf{.861} & .839  & .847  & .853 \\
          & {\textbf{maxE} $\uparrow$} & .717  & .785  & .782  & .736  & .786  & .800  & .700  & .779  & .828  & .698  & .865  & .894 & .807  & .882 & \textbf{.897} & .852  & .870  & .870 \\
          & {\textbf{adpF} $\uparrow$} & .580  & .656  & .749  & .613  & .679  & .693  & .522  & .674  & .724  & .628  & .710  & .791 & .694  & .748  & .767  & .726  & .801 & \textbf{.801} \\
          & {\textbf{meanF} $\uparrow$} & .367  & .469  & .453  & .489  & .572  & .564  & .515  & .470  & .624  & .347  & .689  & .777 & .672  & .721  & .773  & .747  & .786 & \textbf{.791} \\
          & {\textbf{maxF} $\uparrow$} & .568  & .682  & .740  & .619  & .711  & .710  & .535  & .703  & .735  & .592  & .729  & .807  & .687  & .781  & .810 & .766  & \textbf{.812} & .810 \\
          & {\textbf{MAE} $\downarrow$} & .195  & .203  & .180  & .278  & .169  & .165  & .214  & .192  & .142  & .196  & .099  & \textbf{.062}  & .118  & .082  & .063 & .082 & .068  & .066 \\
    \bottomrule
    \end{tabular}%
    }
  \label{table:Quantitative}%
\end{table*}%

\subsection{Performance Comparisons}
\label{Sec:PerformanceComparisons}
We have compared our method with 16 SOTA RGB-D salient object detection methods over all the adopted 5 benchmark datasets.
The adopted SOTA methods include LHM~\cite{ECCV_P2014}, ACSD~\cite{ICIP_J2014}, GP~\cite{CVPRW_J2015}, LBE~\cite{CVPR_F2016}, DCMC~\cite{Cong2016Saliency}, SE~\cite{guo2016salient}, CDCP~\cite{zhu2017innovative}, MDSF~\cite{song2017depth}, DF~\cite{TIP_Q2017}, CDB~\cite{liang2018stereoscopic}, CTMF~\cite{han2017cnns}, PCA~\cite{chen2018progressively}, AFNet~\cite{wang2019adaptive}, MMCI~\cite{chen2019multi}, TANet~\cite{chen2019three} and CPFP~\cite{zhao2019contrast}.
For a fair comparison, the saliency maps/executable codes of the compared methods are all provided by the authors with parameters/implementations unchanged.

\begin{figure}[t]
  \begin{center}
  \includegraphics[width=1\linewidth]{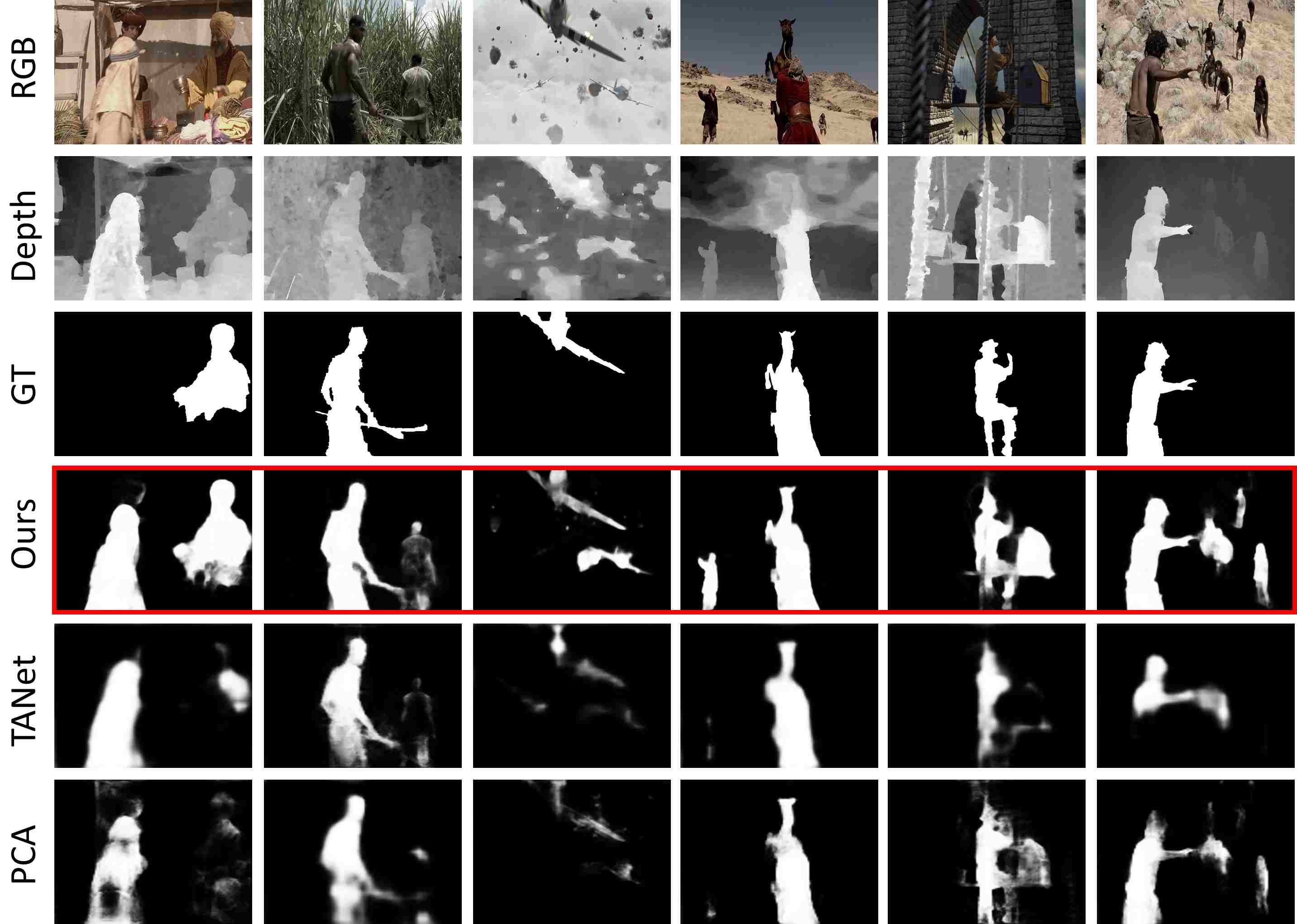}
  \end{center}
  \caption{Demonstration of incomplete human-annotated ground truth masks (GT) in SSD, which are mainly induced by subjective annotations.}
  \label{fig:Demo_SSD}
  \vspace{-0.4cm}
  \end{figure}

\begin{figure*}[t]
  \centering
  \includegraphics[width=1\linewidth]{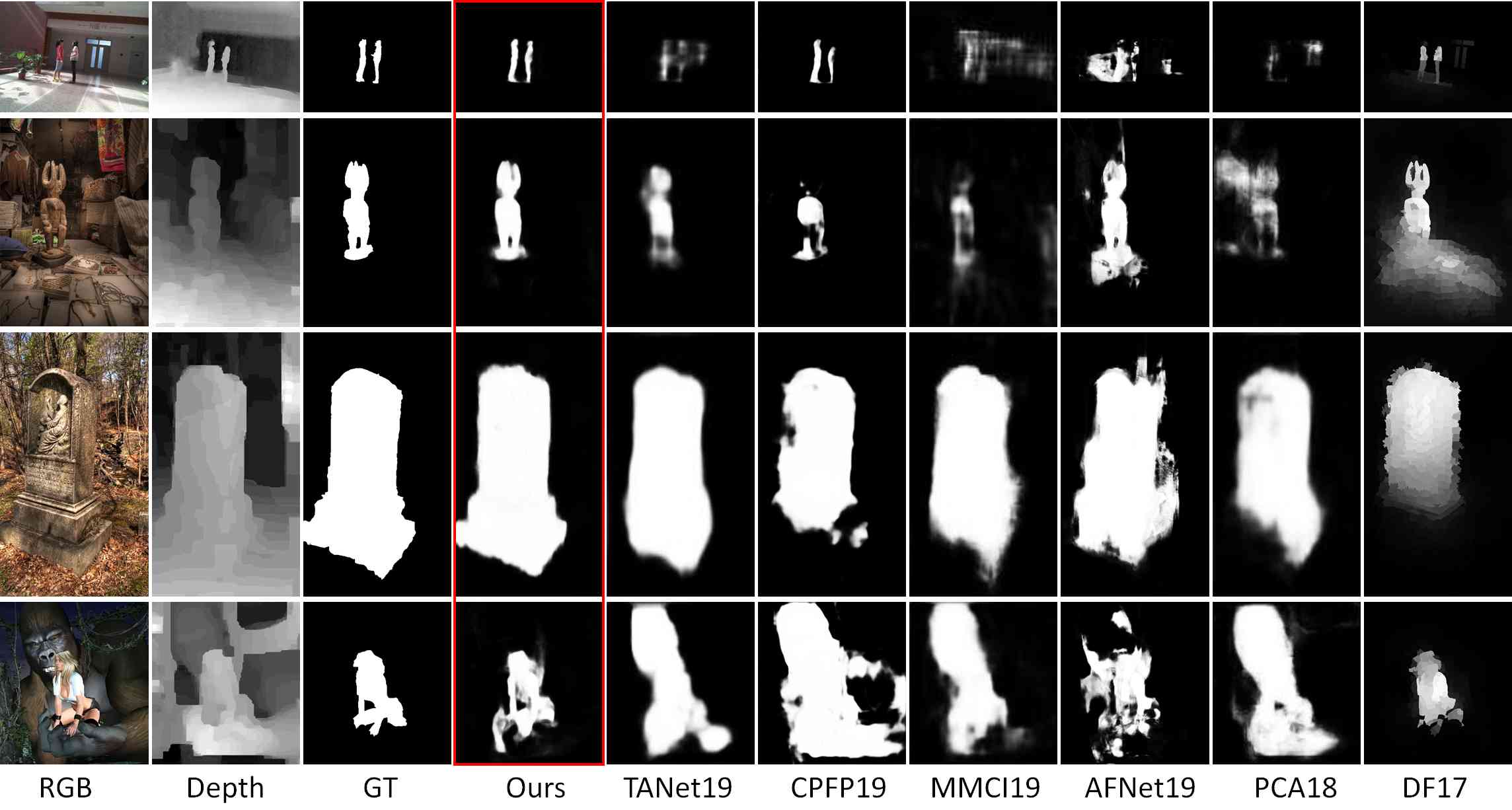} 
  \caption{Qualitative comparisons between our method and other 6 most representative SOTA methods, including the TANet~\cite{chen2019three}, CPFP~\cite{zhao2019contrast}, MMCI~\cite{chen2019multi}, AFNet~\cite{wang2019adaptive}, PCA~\cite{chen2018progressively} and DF~\cite{TIP_Q2017}.}
  \vspace{-0.4cm}
  \label{fig:QuanCom}
\end{figure*}

As shown in Table~\ref{table:Quantitative}, our method can persistently outperform all the compared SOTA methods in terms of 8 metrics over all the 5 adopted benchmark datasets.
Specifically, our method has achieved significant performance improvements over the NJUDS, STERE, DES and NLPR datasets.
The advantage of our method is not much obvious over the SSD dataset due to its controversial human-labeled annotations. Some saliency ground truth annotations in the SSD dataset may be somewhat controversial, i.e., there are totally 80 images in the SSD dataset, and almost 21\% (17/80) of them clearly contain multiple salient objects, while the saliency ground truth annotations of these images tend to regard only one of them as the salient one.
To better understand this issue, we demonstrate several most representative cases in Fig.~\ref{fig:Demo_SSD}, where most of these images contain multiple salient objects, while only one of them is annotated as the salient object.
In these cases, most of the SOTA approaches tend to focus on a single salient object, however, our method tends to detect more objects, and this is exactly why our method is capable of outperforming other approaches in terms of detection completeness over other four datasets. Though facing this problem, our method still belongs to the top-3 methods over the SSD dataset.

We have demonstrated the qualitative comparison results in Fig.~\ref{fig:QuanCom}.
We may easily notice that the salient object detection results of the compared SOTA methods may frequently get degenerated when either the RGB component or the depth part is partially failed to separate the salient object from its non-salient nearby surroundings.
For example, in the bottom row of Fig.~\ref{fig:QuanCom}, the salient object in the RGB component has exhibited a large uniqueness degree, however, it is difficult for the SOTA methods to achieve the correct saliency estimation over the depth channel, leading to an incomplete detection result.

Meanwhile, as shown in the second row of Fig.~\ref{fig:QuanCom}, its depth component is extremely useful in this case, however, almost all the compared methods are incapable of producing correct detection result due to the low contrast RGB component, which leads to an inappropriate complementary status between RGB and depth information.


\vspace{-0.2cm}
\section{Conclusion}
In this paper, we have developed a novel channel-wise fusion network to conduct multi-net and multi-level selective fusion for a high-performance RGB-D salient object detection.
To achieve it, we have newly designed a novel backbone network, which receives our newly formulated input data to pursue an optimal complementary status between RGB and depth channels.
Specifically, our backbone network is implemented based on the VGG16 network, which can be replaced by other high-performance networks, further improving the overall performance of our method.
Then, we have proposed a novel triple-stream fusion network to ensure an optimal fusion state for each of our subnetworks in a multi-level fashion manner.
Moreover, we have conducted extensive quantitative evaluations to verify the effectiveness of our method.

\textbf{Acknowledgments}. This research is supported in part by
National Natural Science Foundation of China (No. 61802215
and No. 61806106), Natural Science Foundation of Shandong
Province (No. ZR2019BF011 and ZR2019QF009) and National Science Foundation of USA (No. IIS-1715985 and IIS-
1812606)


\setcounter{equation}{0}   
\renewcommand{\theequation}{A\arabic{equation}} 
\setcounter{figure}{0}   
\renewcommand{\thefigure}{A\arabic{figure}} 


\ifCLASSOPTIONcaptionsoff
  \newpage
\fi
\bibliographystyle{IEEEtran}
\bibliography{reference}

\end{document}